\documentclass[conference]{IEEEtran}
\IEEEoverridecommandlockouts

\usepackage{cite}
\usepackage{amsmath,amssymb,amsfonts}
\usepackage{graphicx}
\usepackage{textcomp}
\usepackage[table,dvipsnames]{xcolor}
\usepackage{graphics}
\usepackage{graphicx}
\usepackage{threeparttable}
\usepackage{multirow} 

\usepackage{algorithm}
\usepackage{algpseudocode}

\usepackage{mathtools}
\usepackage{amsmath}
\usepackage{acronym}
\usepackage[hidelinks]{hyperref}
\hypersetup{breaklinks=true}
\usepackage{graphicx}
\usepackage{textcomp}
\usepackage{fancyhdr}

\usepackage{siunitx}  

\usepackage{fontawesome}
\usepackage{amsfonts}
\usepackage{pifont}
\usepackage[super]{nth}
\usepackage{tikz}
            
\newcommand*\circled[1]{\tikz[baseline=(char.base)]{ \node[shape=circle,fill=gray,inner sep=0.4pt] (char) {\textcolor{white}{#1}};}}

\usepackage{url}

\def\BibTeX{{\rm B\kern-.05em{\sc i\kern-.025em b}\kern-.08em
    T\kern-.1667em\lower.7ex\hbox{E}\kern-.125emX}}

\usepackage{enumitem}    

\newif\ifoutline
\outlinetrue

\newcommand{\blue}[1]{\ifoutline{\color{black}#1}\fi}
\newcommand{\orange}[1]{\ifoutline{\color{black}#1}\fi}

\renewcommand{\IEEEbibitemsep}{0pt plus 0.5pt}
\makeatletter
\IEEEtriggercmd{\reset@font\normalfont\fontsize{7.9pt}{8.60pt}\selectfont}
\newcommand*\titleheader[1]{\gdef\@titleheader{#1}}
\AtBeginDocument{%
  \let\st@red@title\@title
  \def\@title{%
    \bgroup\normalfont\normalsize\centering\@titleheader\par\egroup
    \vskip1ex\st@red@title}
}
\makeatother
\IEEEtriggeratref{1}

\bstctlcite{IEEEexample:BSTcontrol} 
\author{
    \IEEEauthorblockN{
        Giorgos Armeniakos\IEEEauthorrefmark{1},
        Georgios Mentzos\IEEEauthorrefmark{1},
        Dimitrios Soudris\IEEEauthorrefmark{1}
    }
    \IEEEauthorblockA{
        \IEEEauthorrefmark{1}National Technical University of Athens, GR,
    }
    \IEEEauthorblockA{
        \IEEEauthorrefmark{1}\{armeniakos, gmentzos, dsoudris\}@microlab.ntua.gr    }
}

\titleheader{\vspace{-41pt}\color{gray}To appear at the 31st International Conference
on Electronics Circuits and Systems (ICECS), Nov 18-20 2024, Nancy, France.}
\title{\vspace{-8pt}Accelerating TinyML Inference on Microcontrollers through Approximate Kernels\\
\thanks{\hspace{-2ex}Work partially supported by the Horizon Europe research and innovation program via the “CONVOLVE” project under grant agreement No. 101070374.}
}

\begin{document}

\maketitle

\IEEEpubid{\begin{minipage}{\textwidth}\ \\[5pt]
    \centering
    \color{gray}
    © 2024 IEEE. Personal use of this material is permitted. Permission from IEEE must be obtained for all other uses, in any current or future media, including reprinting/republishing this material for advertising or promotional purposes, creating new collective works, for resale or redistribution to servers or lists, or reuse of any copyrighted component of this work in other works.\vspace{-23ex}
\end{minipage}}

\begin{abstract}

The rapid growth of microcontroller-based IoT devices has opened up numerous applications, from smart manufacturing to personalized healthcare.
Despite the widespread adoption of energy-efficient microcontroller units (MCUs) in the Tiny Machine Learning (TinyML) domain, they still face significant limitations in terms of performance and memory (RAM, Flash).
In this work, we combine approximate computing and software kernel design to accelerate the inference of approximate CNN models on MCUs.
Our kernel-based approximation framework firstly unpacks the operands of each convolution layer and then conducts an offline calculation to determine the significance of each operand.
Subsequently, through a design space exploration, it employs a computation skipping approximation strategy based on the calculated significance.
Our evaluation on an STM32-Nucleo board and 2 popular CNNs trained on the CIFAR-10 dataset shows that, compared to state-of-the-art exact inference, our Pareto optimal solutions \orange{can feature on average 21\% latency reduction with no degradation in Top-1 classification accuracy, while for lower accuracy requirements, the corresponding reduction becomes even more pronounced.}

\end{abstract}

\begin{IEEEkeywords}
Approximate Computing, MCUs, TinyML 
\end{IEEEkeywords}

\section{Introduction}

In recent years, the proliferation of low-cost IoT microcontroller units (MCUs) has significantly expanded the Tiny Machine Learning (TinyML) domain~\cite{tinyml:csur}, enabling real-time data processing on tiny devices. Despite the energy efficiency of MCUs, their limited resources and high latency challenge the deployment of deep learning models on small-scale hardware. Consequently, new optimizations and customized architectures are needed to bridge the resource gap, making reconfigurable MCUs an attractive option for ML acceleration.

In this effor, ARM's CMSIS-NN~\cite{cmsis} software library offers efficient neural network operations for MCUs running on Arm Cortex-M CPUs, achieving nearly an 11x latency improvement compared to TensorFlow Lite Micro on several ImageNet models deployed on an STM32H743 board. TinyEngine~\cite{mcunet}, a system-model co-design framework, combines neural architecture search with a memory-optimized inference library, resulting in average latency and SRAM usage reductions of 2.1× and 2.4×, respectively, compared to CMSIS-NN. However, relevant frameworks focus on fitting models within memory constraints rather than reducing inference latency of large models. For instance, TinyEngine requires about 1.3s to execute an mcunet-in4 ImageNet model on a 160MHz MCU, highlighting existing latency challenges in real-time applications.

In this work we investigate the feasibility of the efficient utilization of MCUs to enhance DNN performance. By integrating Approximate Computing~\cite{arm:csur} (AC) principles with optimized software kernels, we develop an automated framework that generates specialized approximate code for specific Convolutional Neural Networks (CNNs). Our approach utilizes flash memory to unpack kernel code within convolution layers, eliminating instruction overheads. Subsequently, by leveraging the unpacked operations and the fact that each computation contributes uniquely to the final output, we employ an offline significance-aware computation skipping approach, where certain operations are either skipped or retained. Through design space exploration (DSE), our framework identifies Pareto-optimal solutions, each offering unique accuracy-latency trade-offs tailored to user requirements.
Compared to the state-of-the-art CMSIS-NN, our approach achieves a 21\% latency reduction with no degradation in Top-1 classification accuracy on CIFAR-10 trained CNN models, while for lower accuracy requirements ($<$ 5\%), our method outperforms even commercial frameworks.

\textbf{Our novel contributions within this work are as follows:}
\begin{enumerate}[topsep=0pt,leftmargin=*]
    \item This is the first work that evaluates the impact of approximate computing on the optimized inference library of CMSIS-NN, targeting MCUs.
    \item We propose an automated cooperative approximation framework for accelerating CNNs inference on MCUs\footnote{Available at \href{https://github.com/GeorgeMentzos/ATAMAN-AuTo-driven-Approximation-and-Microcontroller-AcceleratioN-Toolkit}{https://github.com/GeorgeMentzos/ATAMAN-AuTo-driven-Approximation-and-Microcontroller-AcceleratioN-Toolkit}}
    \item Using our framework, we demonstrate that, in many cases approximate computing is able to realize larger and faster networks than conventional ones on tiny devices.
\end{enumerate}

\section{Cooperative Approximation Framework for Inference Optimization}
\begin{figure}
    \centering
    \includegraphics{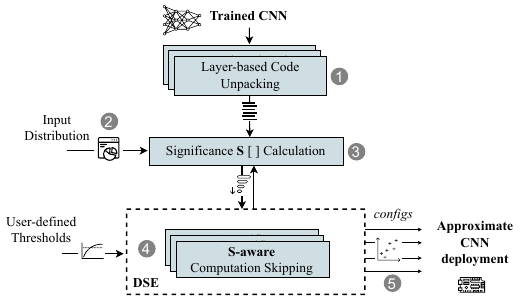}
    \caption{Abstract overview of our framework}
    \label{fig:overview}\vspace{-2ex}
\end{figure}

This section describes our cooperative approximation framework for deploying approximate DNNs on microcontrollers.
In brief, we first describe our basic kernel customizations and how we eliminate associated overheads from existing inference libraries.
Then, we analyze our layer-based code unpacking showing the latency benefits over typical implementation for the targeted kernels and we finally describe our significance-aware computation skipping exploration that offers the flexibility to trade classification accuracy for further inference acceleration.
An abstract overview of our framework is depicted in Fig.~\ref{fig:overview}

\subsection{Customized kernels for NN deployment}

To meet our deployment scenarios' unique requirements, we use CMSIS-NN as our baseline inference library. Our approximation framework customizes the generated code to support only essential layers and functions for the given model. Unlike CMSIS-NN and most existing inference libraries (e.g., TF-Lite Micro), we offload model structure parameter operations from runtime to compile time, enhancing inference efficiency and reducing flash memory usage by up to 30\%. This extra flash memory allows us to unpack more kernels or entire layers, improving the granularity of our skipping approximation.

Our focus is on optimizing convolutional layers, as most cycles in CNN models are consumed by these operations~\cite{cfu:google}. A convolution operation in CMSIS-NN involves computing a dot product between filter weights and a small receptive field within the input feature map, followed by matrix multiplication. We extend these kernels with cycle counters to profile parts of the C code for individual operators, providing insights into the model's baseline performance. These counters are deactivated during runtime.

Table I shows the characteristics (baseline accuracy, topology, latency) for our models deployed on an STM32-Nucleo-U575ZI-Q board at 160MHz, trained on the CIFAR-10 dataset with 8-bit post-training quantization. Inputs have a 32x32 resolution and are normalized to [0,1]. As shown, even for a small model with less than 5M parameters, latency exceeds 80ms, while for larger models like AlexNet, 87\% of the flash memory remains unused. This inspires us to leverage available flash for customized kernels optimized for specific models.

\begin{table}[t]
\setlength\tabcolsep{5.3pt}
\renewcommand{\arraystretch}{1.3}
\caption{Evaluation of our baseline CIFAR-10 AlexNet and Lenet on a STM32-Nucleo fitting~\blue{2000}KB ROM and~\blue{768}KB RAM }
\begin{threeparttable}
\begin{tabular}{c|c|c|c|c|c|c}
\hline
\rowcolor[HTML]{EFEFEF} 
\multicolumn{1}{c|}{\cellcolor[HTML]{EFEFEF}\textbf{CNN}} &
  \textbf{Acc} &
  \textbf{Topol.\tnote{1}} &
  \textbf{\begin{tabular}[c]{@{}c@{}}\# MAC \\ Ops\end{tabular}} &
  \textbf{\begin{tabular}[c]{@{}c@{}}Latency\\ (ms)\end{tabular}} &
  \textbf{\begin{tabular}[c]{@{}c@{}}Flash \\ Usage (\%)\end{tabular}} &
  \cellcolor[HTML]{EFEFEF}\textbf{\begin{tabular}[c]{@{}c@{}}RAM\\ (\blue{KB})\end{tabular}} \\ \hline
\textbf{AlexNet} &
  71.9 &
  5-2-2 &
  \orange{16.1}M &
  179.9 &
  13 &
  212.16 \\
\rowcolor[HTML]{EFEFEF} 
\textbf{LeNet} &
  \orange{71.6} &
  3-2-2 &
  \orange{4.5} M &
  82.8 &
  12 &
  \cellcolor[HTML]{EFEFEF}183.5 \\ \hline
\end{tabular}
\begin{tablenotes}\scriptsize
\item[] $^1$ Topology of network in Conv - MaxPooling - Full Connected layers, respectively. 
\end{tablenotes}
\end{threeparttable}
\label{tab:baseline}\vspace{-3ex}
\end{table}

\subsection{Layer-based code unpacking}\label{sec:unpack}

Typical convolution kernels on MCUs are usually implemented in a matrix format, where inputs and weights are retrieved from memory using a specific pattern.
This pattern includes details such as the order in which data elements are fetched, the stride or step size for moving through the data and any necessary padding or adjustments to ensure the correct alignment of the data for convolution.
Instead, our framework performs an automated layer-based code unpacking (see Fig.~\ref{fig:overview} \circled{1}), where each operation is ``unpacked'' and included as an intrinsic function in the final generated code.
Our unpacking technique fundamentally differs from typical unrolling, since it utilizes known constant values (weights) within each iteration. This approach enables more optimized and efficient code generation, as it allows for additional compiler optimizations.
\orange{The primary benefits of our code unpacked kernels that lead to reduced execution cycles, include the followings:
\begin{enumerate}[leftmargin=*]
\item Similar to typical unrolling techniques, our code unpacking also \textbf{eliminates} branch instruction overheads within convolutional kernels.
\item Our automated procedure allocates fixed weights to each operand, \textbf{excluding} the necessity to adapt and load the weights properly during the convolution process. 
This leads to simplified and predictable, in terms of type, operations that can be adjusted based on input values to enhance inference speed.
\item CMSIS $mat\_mult$ kernel calculates the partial products using the SMLAD instruction (SIMD logic), which performs two 16-bit signed multiplications, accumulating the results into a 32-bit operand.
Hence, a pre-processing is required to convert the data to the 16-bit data type.
Instead, our fixed-weight replacement \textbf{avoids} this time-consuming operation.
Since we know apriori the values of weights, this is easily avoided by an offline processing that involves concatenating two int16 (sign-extended int8 values to int16) weights.
As an instance, an SMLAD (MAC) instruction with the ``hardwired'' value of $w_{12}$=4194324 represents two multiplications with $w_1$=64 and $w_2$=20, as $64\cdot2^{16} + 20 = 4194324$.
\end{enumerate}
}
The length of the unpacked code is considered with respect to the available unused flash memory, creating an interesting trade-off between these two metrics.
Along with works like~\cite{mcunet} and~\cite{tinymlcontest} that report an average flash memory utilization of less than 25\%, we demonstrate in this work that a fully unpacked fixed-weight convolution can be effortlessly enabled.
For instance, even in the worst case of AlexNet with~\blue{5} convolution layers, our framework fitted the whole kernel instructions using less than 60\% of the available flash memory.

\subsection{Significance-aware skipping}\label{sec:skipping}
In this section, we describe how we leverage our layer-based code unpacking to systematically omit certain operations that are considered \emph{insignificant} for our classification tasks or choose to retain as \emph{significant} some others. 
Unlike other approaches that consider skipping entire channels or even layers~\cite{insignif:isca}, our framework can omit operations at the finest granularity, which, to the best of our knowledge, no other work has targeted before in software libraries for MCUs.
Our significance-driven analysis is motivated by two facts: 1) each computation within the convolution makes a unique contribution to the final output, meaning that certain computations could potentially be skipped without compromising classification accuracy and 2) effectively reducing the total number of computations could provide a valuable trade-off between accuracy and latency.



The accumulation of each channel during the matrix multiplication (\textit{mat\_mult} kernel) is calculated based on a weighted sum and an initialized bias:\\
\begin{equation}\label{eq:sum}
    Sum_c = b + \sum_{\forall i}{a_i\cdot w_i},
\end{equation}
where, $b$ is the initialized bias, $w_i$ are the trained coefficients (weights) and $a_i$ are the inputs from the respective channel.
Intuitively, when inputs are multiplied by large numbers, they tend to produce significantly more impactful products ($a_i\cdot w_i$) in the final result compared to inputs multiplied by small values. 
However, it is worth noting that the significance of the product $a_i\cdot w_i$ also depends on the value of $a_i$.
Thus, we define the \emph{significance} (\circled3) of each product as follows:

\begin{equation}\label{eq:sig}
    S_i=|\frac{\mathrm{E}[a_i] \cdot w_i}{\sum_{\forall i}{\big(\mathrm{E}[a_i]\cdot w_i \big)}}|
\end{equation}

, \orange{where $E[a_i]$ is the average expected value of the $a_i$ input.
In other words,~\eqref{eq:sig} calculates the long-term expected outcome of each product $a_i\cdot w_i$ over the total sum $Sum_c$ of the respective channel.
If the sum equals with zero, which is the vast minority of the cases, we consider the corresponding significance $S_i$ to be large, and thus, the product is retained.}
For each channel and $Sum_c$, the calculation of $S_i, \forall i$, is straightforward and involves capturing the input values' distribution (\circled2) from a small portion of the dataset.

By exploiting this high-level information, we minimize the total computations required for each summation ($Sum_c$) at compile time, and thus, we approximate the summation accordingly.
Specifically, for each product $a_i \cdot w_i$, if $S_i$ is less or equal to a given threshold
$\tau$ , it is incorporated into the generated code, while others are omitted.
Thus, our approximate summation per channel is now represented by:
\begin{equation}
\begin{gathered}\label{eq:axwsumk}
Sum_c^\prime=b + \sum_{\forall i}{(a_i\cdot w_i)} - \sum_{\mathclap{\substack{\forall i: S_i\leq\tau}}}{(a_i \cdot w_i)}\,
\end{gathered}
\end{equation}

Finally, we perform an exhaustive DSE w.r.t. the targeted layers and the values of $\tau$ ranging from~\blue{[0, \orange{0.1}] with a step of \orange{0.001} and 0.01 for LeNet and AlexNet, respectively.}
This exploration is performed offline and only once.
Every approximate configuration, denoting which layers and computations are approximated, undergoes simulation to calculate the classification accuracy.
Subsequently, a Pareto analysis is conducted to determine trade-offs between accuracy and total perforated MAC operations, leading to a model with increased speedup.
Note that in this work, we exclusively concentrate on the convolution layers, and therefore, the model's behavior, when considering the rest of the functions, becomes rather predictable. 
Consequently, the clock cycles reported by our counters during our simulations~\cite{cfu:google} closely align with the cycles of the actual model deployment, and export representative gain percentages with respect to the ``unpacked'' model.
On average, DSE required less than \blue{2} hours using \blue{6} threads.
The aforementioned execution times refer to an \blue{Intel-i7-8750H with 32GB RAM}.
Following the DSE analysis, we extract the suitable approximate configuration (\circled5) based on the user's specified accuracy loss threshold and desired possible speedup.
Subsequently, our framework generates the approximate code (\circled4), which is then compiled and deployed to the MCU.

\section{Experimental Results and Analysis}

In this section we evaluate the efficiency of our proposed framework in reducing inference latency at the cost of some classification accuracy and we investigate the impact of approximate computing within the context of TinyML on MCUs.
We evaluate the inference latency, classification accuracy, memory usage and energy of our approximate designs against the state-of-the-art exact models~\cite{cmsis} and we also compare our framework against the closed-source X-CUBE-AI~\cite{xcubeai} framework.
All the experiments are evaluated on an STM32\blue{U575ZIT6Q} SoC, an ARM Cortex-M33 based MCU, running at 160 MHz, with 2 MB of Flash and \blue{768}KB of RAM.


\begin{table*}[t]
\setlength\tabcolsep{3.6pt}
\renewcommand{\arraystretch}{1.1}
\caption{Comparison with state-of-the-art CMSIS~\cite{cmsis} and X-CUBE-AI~\cite{xcubeai} for two CNNs deployed on an \blue{stm32u575zi-q} board fitting \blue{2MB} Flash and 768KB RAM. Three accuracy loss thresholds have been considered.}
\begin{threeparttable}
\begin{tabular}{l|cc|cc|cccccc}
\hline
\textbf{} &
  \multicolumn{2}{c|}{\textbf{CMSIS-NN}} &
  \multicolumn{2}{c|}{\textbf{X-CUBE-AI}} &
  \multicolumn{6}{c}{\textbf{Proposed (ours)}} \\ \hline \hline
\textbf{Network} &
  \textbf{LeNet} &
  \textbf{AlexNet} &
  \textbf{LeNet} &
  \textbf{AlexNet} &
  \textbf{LeNet(\orange{0}\%)} &
  \textbf{LeNet(\orange{5}\%)} &
  \multicolumn{1}{c|}{\textbf{LeNet(\orange{10}\%)}} &
  \textbf{AlexNet(\orange{0}\%)} &
  \textbf{AlexNet(\orange{5}\%)} &
  \textbf{AlexNet(\orange{10}\%)} \\ \cline{2-11} 


\textbf{Top-1 Accuracy (\%)}& \orange{\textbf{71.6}} & \textbf{71.9} & \orange{71.6}& 71.9 & \orange{71.6} & \orange{66.7} & \orange{61.6} &  \orange{72.4} & \orange{67.1} & \orange{62.1}\\
\textbf{Latency (ms)} & 82.8  & 179.9 & \textbf{63.5} & 150.7 & \orange{72.7}  & \orange{66.8} & \orange{59.8} & \orange{\textbf{124.8}} & \textbf{\orange{111.3}} & \textbf{\orange{101.5}}\\ \hline
\textbf{Flash (KB)}   & 239   & 267   & 154 & 178 & \orange{761}   & \orange{704}  & \orange{681}  & \orange{1080}  & \orange{954}  & \orange{891}\\
\textbf{\#MAC Ops.}  & \orange{4.5M} & \orange{16.1M} & \orange{4.5M} & \orange{16.1M} & \orange{\textbf{3.3M}} & \textbf{\orange{2.9M}} & \textbf{\orange{2.4M}} & \textbf{\orange{7.5M}} & \textbf{\orange{6.2M}} & \textbf{\orange{5.5M}} \\
\textbf{Energy (mJ)} & 2.73 & 5.94 & \textbf{2.10} & 4.97 & \orange{2.40} & \orange{2.20} & \orange{1.98} & \orange{\textbf{4.12}} & \textbf{\orange{3.67}} & \textbf{\orange{3.35}} \\ \hline
\end{tabular}

\end{threeparttable}
\label{tab:results}\vspace{-3ex}
\end{table*}

\begin{figure}
    \centering
    \includegraphics[scale=0.70]{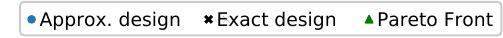}
    \includegraphics[scale=0.65]{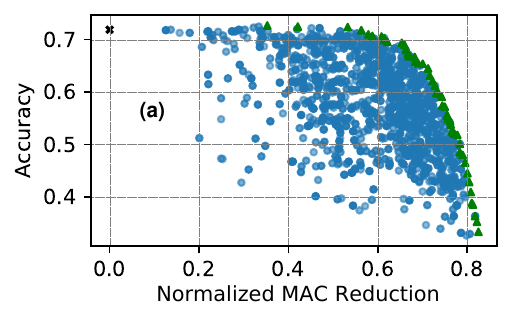}
    \includegraphics[scale=0.65]{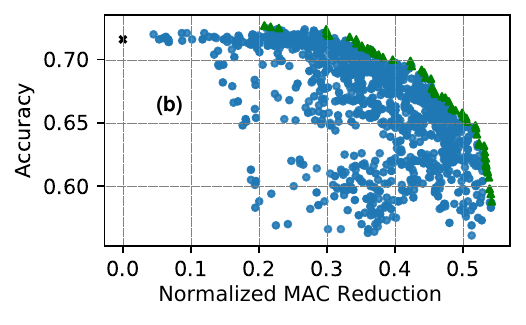}
    \caption{\orange{Pareto space between accuracy and normalized MAC unit reduction is depicted for our computation skipping approach within all convolution layers for AlexNet (a) and LeNet (b).}}
    \label{fig:dse} \vspace{-3ex}
\end{figure}


\blue{Before deploying the final approximate design and measuring its latency, an initial analysis is required (\circled{5}).
This offline analysis assists in extracting approximate designs based on the specified accuracy loss threshold set by the user and also avoids reconfiguring the MCU multiple times (i.e., $\equiv$~designs of DSE), which could potentially result in flash memory deterioration.
Hence, Fig.~\ref{fig:dse} presents the Pareto space between accuracy and normalized MAC unit reduction achieved from our skipping approximation for the two examined CNNs.
In Fig.~\ref{fig:dse} MAC reduction concerns only the convolution layers. 
}
The black `x' is our exact baseline design~\cite{cmsis}.
The blue dots on the graph correspond to approximate configurations. 
The green triangles, on the other hand, form the Pareto Front line.
These configurations particularly represent the percentage of operations that are skipped and the indexes of these operations in the final generated code.
Note that the number of the explored configuration/designs is model dependent.
As aforementioned, the DSE was performed based on various significance thresholds $\tau$, steps and examined layers for both CNNs.
In total, we evaluated more than \orange{10,000} approximate designs for LeNet and AlexNet, separately.
On average, it is observed that our ``only skipping'' approximation achieves \orange{44\%} \orange{MAC} reduction, delivering identical classification accuracy with the exact baseline, while this number rises further for both models to averagely \orange{57\%} when compromising \orange{5\%} accuracy loss.


In Table~\ref{tab:results} we report some important metrics of our framework.
To generate this table we considered three conservative accuracy loss thresholds (i.e., \orange{0\%, 5\% and 10\%}) and we report the latency, Top-1 accuracy, flash, and energy metrics for the latency-optimized approximate designs after deployment on the examined MCU.
As aforementioned, due to the nature of target AI applications, such as real-time processing, a fast inference is one of the foremost requirements when targeting DNNs on MCUs and so prioritizing it over strict accuracy constraints is a typical procedure~\cite{mcunet}.
As depicted in Table~\ref{tab:results}, our cooperative approximation approach, which includes both code unpacking and significance-aware skipping approximation, achieves an average a speedup of \orange{21}\% while incurring no degradation (zero accuracy loss) compared to the exact baseline~\cite{cmsis}. 
Moreover, the respective speedup is increased to \orange{36}\% when accepting approximately \orange{10}\% accuracy loss.
In Table~\ref{tab:results}, we also provide a comparison of our models with the state-of-the-art homogeneous inference library, X-CUBE-AI. 
Although X-CUBE-AI attains a \orange{12}\% lower latency for the precise LeNet(0\%) compared to our framework, it is worth noting that for the more complex CNN of AlexNet, our approach outperforms X-CUBE-AI.
Specifically, we achieve an increased speedup of \orange{17}\% with identical classification accuracy, while even for LeNet we manage better latency for \orange{7}\% accuracy loss.
As shown, our framework, can surpass even commercial tools like X-CUBE-AI (that have also very limited flexibility), providing an accuracy-latency trade-off that was previously unattainable for optimized libraries like CMSIS.


Lastly, we undertake a qualitative evaluation, comparing our approximation framework with other state-of-the-art methodologies.
When compared to CMix-NN~\cite{cmix} using a model with 13.8M MAC operations, our framework achieves a latency of \orange{124}ms on a 160MHz MCU.
This means that, compared to CMix-NN~\cite{cmix}, our framework achieves a remarkable \orange{62}\% reduction in latency, with a negligible accuracy degradation.
Additionally, uTVM~\cite{microtvm}, an end-to-end ML compiler framework tailored for bare-metal MCUs, reports a 13\% latency overhead compared to CMSIS when using a similar LeNet model architecture.
For the same model, our approach outperforms uTVM, achieving an additional \orange{32}\% speedup with an accuracy loss of less than \orange{5}\%.


\section{Conclusion}

In this work, to address the notable latency limitations of MCUs, we introduce a cooperative framework that combines approximate computing with software kernel optimizations. 
Through a systematic kernel-based computation skipping approach, our framework effectively removes operations deemed insignificant for the model's inference, resulting in accelerated inference speeds at the expense of different accuracy trade-off.
These trade-offs have the potential to open avenues for more AI applications and enable the execution of more complex deep neural networks on tiny MCUs.



\makeatletter
\def\footnoterule{\kern-3\p@
  \hrule \@width 0.75in \kern 2.6\p@} 
\makeatother


{
\linespread{0.94}\selectfont
\bibliographystyle{IEEEtran}
\bibliography{references}

\begin{thebibliography}{10}
\providecommand{\url}[1]{#1}
\csname url@samestyle\endcsname
\providecommand{\newblock}{\relax}
\providecommand{\bibinfo}[2]{#2}
\providecommand{\BIBentrySTDinterwordspacing}{\spaceskip=0pt\relax}
\providecommand{\BIBentryALTinterwordstretchfactor}{4}
\providecommand{\BIBentryALTinterwordspacing}{\spaceskip=\fontdimen2\font plus
\BIBentryALTinterwordstretchfactor\fontdimen3\font minus \fontdimen4\font\relax}
\providecommand{\BIBforeignlanguage}[2]{{%
\expandafter\ifx\csname l@#1\endcsname\relax
\typeout{** WARNING: IEEEtran.bst: No hyphenation pattern has been}%
\typeout{** loaded for the language `#1'. Using the pattern for}%
\typeout{** the default language instead.}%
\else
\language=\csname l@#1\endcsname
\fi
#2}}
\providecommand{\BIBdecl}{\relax}
\BIBdecl

\bibitem{tinyml:csur}
V.~Rajapakse, I.~Karunanayake, and N.~Ahmed, ``Intelligence at the extreme edge: A survey on reformable tinyml,'' \emph{ACM Comput. Surv.}, vol.~55, no. 13s, jul 2023.

\bibitem{cmsis}
L.~Lai, N.~Suda, and V.~Chandra, ``Cmsis-nn: Efficient neural network kernels for arm cortex-m cpus,'' 01 2018.

\bibitem{mcunet}
J.~Lin, W.-M. Chen, Y.~Lin, J.~Cohn, C.~Gan, and S.~Han, ``Mcunet: Tiny deep learning on iot devices,'' in \emph{Proceedings of the 34th International Conference on Neural Information Processing Systems}, ser. NIPS'20.\hskip 1em plus 0.5em minus 0.4em\relax Red Hook, NY, USA: Curran Associates Inc., 2020.

\bibitem{arm:csur}
G.~Armeniakos, G.~Zervakis, D.~Soudris, and J.~Henkel, ``Hardware approximate techniques for deep neural network accelerators: A survey,'' \emph{ACM Comput. Surv.}, vol.~55, no.~4, nov 2022.

\bibitem{cfu:google}
S.~Prakash, T.~Callahan, J.~Bushagour, C.~Banbury, A.~V. Green, P.~Warden, T.~Ansell, and V.~J. Reddi, ``Cfu playground: Full-stack open-source framework for tiny machine learning (tinyml) acceleration on fpgas,'' in \emph{2023 IEEE International Symposium on Performance Analysis of Systems and Software (ISPASS)}, 2023, pp. 157--167.

\bibitem{tinymlcontest}
Z.~Jia, D.~Li, C.~Liu, L.~Liao, X.~Xu, L.~Ping, and Y.~Shi, ``Tinyml design contest for life-threatening ventricular arrhythmia detection,'' \emph{IEEE Transactions on Computer-Aided Design of Integrated Circuits and Systems}, pp. 1--1, 2023.

\bibitem{insignif:isca}
J.~Zhang, X.~Chen, M.~Song, and T.~Li, ``Eager pruning: Algorithm and architecture support for fast training of deep neural networks,'' in \emph{2019 ACM/IEEE 46th Annual International Symposium on Computer Architecture (ISCA)}, 2019, pp. 292--303.

\bibitem{xcubeai}
\BIBentryALTinterwordspacing
STMicroelectronics, ``X-cube-ai: Ai expansion pack for stm32cubemx,'' Dec 2019. [Online]. Available: \url{https://www.st.com/en/embedded-software/x-cube-ai.html}
\BIBentrySTDinterwordspacing

\bibitem{cmix}
A.~Capotondi, M.~Rusci, M.~Fariselli, and L.~Benini, ``Cmix-nn: Mixed low-precision cnn library for memory-constrained edge devices,'' \emph{IEEE Transactions on Circuits and Systems II: Express Briefs}, vol.~67, 2020.

\bibitem{microtvm}
T.~Chen, T.~Moreau, Z.~Jiang, L.~Zheng, E.~Yan, M.~Cowan, H.~Shen, L.~Wang, Y.~Hu, L.~Ceze, C.~Guestrin, and A.~Krishnamurthy, ``Tvm: An automated end-to-end optimizing compiler for deep learning,'' in \emph{Proceedings of the 13th USENIX Conference on Operating Systems Design and Implementation}, ser. OSDI'18, 2018.

\end{thebibliography}
}

\end{document}